\documentclass[10pt,twocolumn,letterpaper]{article}

\usepackage{cvpr}              % To produce the CAMERA-READY version
\usepackage{graphicx}
\usepackage{amsmath}
\usepackage{amssymb}
\usepackage{booktabs}
\usepackage{xcolor}
\usepackage{multirow}

\usepackage[pagebackref,breaklinks,colorlinks]{hyperref}

\usepackage[capitalize]{cleveref}
\Crefname{section}{Sec.}{Secs.}
\Crefname{section}{Section}{Sections}
\Crefname{table}{Table}{Tables}
\Crefname{table}{Tab.}{Tabs.}

\def\cvprPaperID{8462} % *** Enter the CVPR Paper ID here
\def\confName{CVPR}
\def\confYear{2022}

\newcommand{\norm}[1]{\left\lVert#1\right\rVert}

\begin{document}

\title{Hyperbolic Vision Transformers: Combining Improvements in Metric Learning}

\author{Aleksandr Ermolov\\
University of Trento, Italy\\
{\tt\small aleksandr.ermolov@unitn.it}
\and
Leyla Mirvakhabova\\
Skoltech\thanks{Skolkovo Institute of Science and Technology}, Russia\\
{\tt\small leyla.mirvakhabova@skoltech.ru}
\and
Valentin Khrulkov\\
Yandex \& Skoltech\footnotemark[1], Russia\\
{\tt\small khrulkov.v@gmail.com}
\and
Nicu Sebe\\
University of Trento, Italy\\
{\tt\small niculae.sebe@unitn.it}
\and
Ivan Oseledets\\
Skoltech\footnotemark[1] \& AIRI, Russia\\
{\tt\small I.Oseledets@skoltech.ru}
% For a paper whose authors are all at the same institution,
% omit the following lines up until the closing ``}''.
% Additional authors and addresses can be added with ``\and'',
% just like the second author.
% To save space, use either the email address or home page, not both
% \and
% Second Author\\
% Institution2\\
% First line of institution2 address\\
% {\tt\small secondauthor@i2.org}
}
\maketitle

\begin{abstract}
% In this work, we consider the problem of metric learning, defined as a distance between objects representing their semantic similarity. In most cases, there is an encoder, which provides an embedding from an image, and the distance is calculated between such embeddings. We employ a vision transformer as an encoder and consider several pretraining schemes, including a self-supervised learning method. Usually, some form of the Euclidean distance is used as the metric distance. However, several recent works have demonstrated that hyperbolic space is more suitable for natural data. Therefore, we project the embeddings and calculate the distance in the hyperbolic space. Finally, we employ the popular pairwise cross-entropy loss to optimise these distances. We evaluate six versions of our method on four datasets achieving the new state-of-the-art. The source code of the method and of all the experiments is available in the supplementary material.
Metric learning aims to learn a highly discriminative model encouraging the embeddings of similar classes to be close in the chosen metrics and pushed apart for dissimilar ones. The common recipe is to use an encoder to extract embeddings and a distance-based loss function to match the representations -- usually, the Euclidean distance is utilized. An emerging interest in learning hyperbolic data embeddings suggests that hyperbolic geometry can be beneficial for natural data. Following this line of work, we propose a new hyperbolic-based model for metric learning. At the core of our method is a vision transformer with output embeddings mapped to hyperbolic space. These embeddings are directly optimized using modified pairwise cross-entropy loss. We evaluate the proposed model with six different formulations on four datasets achieving the new state-of-the-art performance. The source code is available at \url{https://github.com/htdt/hyp_metric}.
\end{abstract}

\section{Introduction}
\label{sec:intro}

\begin{figure*}
\begin{center}
\includegraphics[width=1\linewidth]{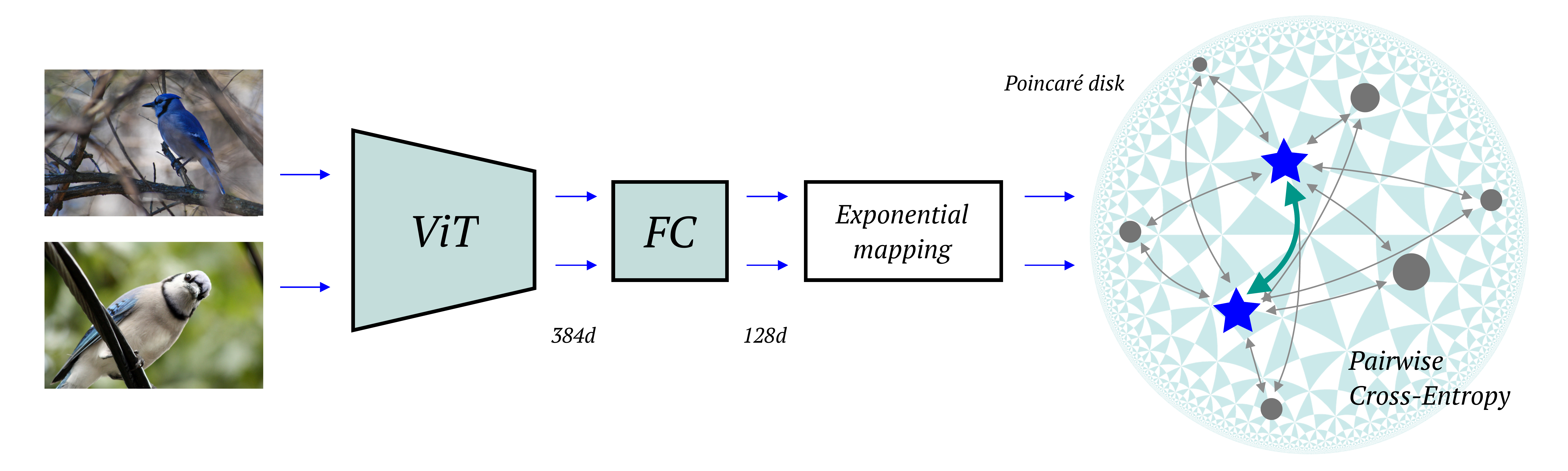}
% \vspace{-2mm}
\caption{Overview of the proposed method. Two images representing one class (positives) are encoded with the vision transformer, projected into a space of a lower dimension with a fully connected (FC) layer, and then mapped to a hyperbolic space. Blue stars depict the resulting embeddings. Poincar\'e disk is shown with uniform triangle tiling on the background to illustrate the manifold curvature. Gray circles represent other samples from the batch (negatives). Finally, arrows in the disk represent distances used in the pairwise cross-entropy loss. Positives are pushed closer to each other, negative are pulled far apart.}
\label{fig:scheme}
\end{center}
\end{figure*}

Metric learning task formulation is general and intuitive: the obtained distances between data embeddings must represent semantic similarity. It is a typical cognitive task to generalize similarity for new objects given some examples of similar and dissimilar pairs. Metric learning algorithms are widely applied in various computer vision tasks: content-based image retrieval \cite{NPair,ProxyNCA,sop}, near-duplicate detection \cite{zheng2016improving}, face recognition \cite{sphereface,facenet}, person re-identification \cite{chen2017triplet,xiao2017joint}, as a part of zero-shot \cite{sop} or few-shot learning \cite{qiao2019transductive,snell2017prototypical,sung2018learning}.
% It also can be a part of zero-shot \cite{sop} or few-shot learning \cite{qiao2019transductive,snell2017prototypical,sung2018learning}.

Modern image retrieval methods can be decomposed into roughly two components: the encoder mapping the image to its compact representation and the loss function governing the training process. Encoders with backbones based on transformer architecture have been recently proposed as a competitive alternative to previously used convolutional neural networks (CNNs). Transformers lack some of CNN's inductive biases, \eg, translation equivariance, requiring more training data to achieve a fair generalization. On the other hand, it allows transformers to produce more general features, which presumably can be more beneficial for image retrieval \cite{IRT,dino}, as this task requires generalization to unseen classes of images. To alleviate the issue above, several training schemes have been proposed: using a large dataset \cite{vit}, heavily augmenting training dataset and using distillation \cite{deit}, using self-supervised learning scenario \cite{dino}. 
%We employ a vision transformer as the encoder for our method investigating these pretraining schemes.

The choice of the embedding space directly influences the metrics used for comparing representations. Typically, embeddings are arranged on a hypersphere, \ie the output of the encoder is $L_{2}$ normalized, resulting in using cosine similarity as a distance. In this work, we propose to consider the hyperbolic spaces. Their distinctive property is the exponential volume growth with respect to the radius, unlike Euclidean spaces with polynomial growth. This feature makes hyperbolic space especially suitable for embedding tree-like data due to increased representation power. The paper \cite{sarkar2011low} shows that a tree can be embedded to Poincar\'e disk with an arbitrarily low distortion. Most of the natural data is intrinsically hierarchical, and hyperbolic spaces suit well for its representation. Another desirable property of hyperbolic spaces is the ability to use low-dimensional manifolds for embeddings without sacrificing the model accuracy and its representation power \cite{nickel2017poincare}. 

The goal of the loss function is straightforward: we want to group the representations of similar objects in the embedding space while pulling away representations of dissimilar objects. Most loss functions can be divided into two categories: proxy-based and pair-based \cite{proxy_anchor}. Additionally to the network parameters, the first type of losses trains proxies, which represent subsets of the dataset \cite{ProxyNCA}. This procedure can be seen from a perspective of a simple classification task: we train matching embeddings, which would classify each subset \cite{reality_check}. At the same time, pair-based losses operate directly on the embeddings. The advantage of pair-based losses is that they can account for the fine-grained interactions of individual samples. Such losses do not require data labels: it is sufficient to have pair-based relationships. This property is crucial for a widely used pairwise cross-entropy loss in self-supervised learning scenario \cite{CPC,moco,simclr}. Instead of labels, the supervision comes from a pretext task, which defines positive and negative pairs. Inspired by these works, we adopt pairwise cross-entropy loss for our experiments.

The main contributions of our paper are the following:
\begin{itemize}
    % \item We adopt vision transformers for image retrieval and evaluate the performance of several pretraining schemes.
    \item We propose to project embeddings to the Poincar\'e ball and to use the pairwise cross-entropy loss with hyperbolic distances. Through extensive experiments, we demonstrate that the hyperbolic counterpart outperforms the Euclidean setting.
    \item We show that the joint usage of vision transformers, hyperbolic embeddings, and pairwise cross-entropy loss provides the best performance for the image retrieval task.
\end{itemize}

\section{Method}

\label{sec:method}
% We propose a new metric learning loss that combines representative expressiveness of the hyperbolic space and the simplicity and generality of the cross-entropy loss. 
% was mentioned before
% We employ hyperbolic distances \TODO{correct?} between samples as logits, thus embeddings of two samples from one category ({\em positives}) are pulled to each other in hyperbolic space, while other samples in the batch ({\em negatives}) are pushed away. Additionally, we include a version with $L_{2}$ normalised embeddings; the distance, in this case, is implemented with cosine similarity.
% We adopt image transformers as the encoder for images. The schematic overview of the proposed method is depicted at \Cref{fig:scheme}. In \Cref{sec:vit} we briefly describe the architecture, pretraining schemes, and the way to project \TODO{correct?} the final embedding into the hyperbolic space.
We propose a new metric learning loss that combines representative expressiveness of the hyperbolic space and the simplicity and generality of the cross-entropy loss. The suggested loss operates in the hyperbolic space encouraging the representatives of one class ({\em positives}) to be closer while pushing the samples from other categories ({\em negatives}) away. 
% For a fair comparison, in our experiments, we include a version with $L_{2}$ normalized embeddings with loss based on cosine similarity.

The schematic overview of the proposed method is depicted at \Cref{fig:scheme}. The remainder of the section is organized as follows. We start with providing the necessary preliminaries on hyperbolic spaces in \Cref{sec:hyperb}, then we discuss the loss function in \Cref{sec:xent} and, finally, we briefly describe the architecture and discuss pretraining schemes in \Cref{sec:vit}.
% We also discuss the curvature of the hyperbolic spaces in \Cref{sec:delta} and review a recent improvement 

\subsection{Hyperbolic Embeddings}
\label{sec:hyperb}
Formally, the $n$-dimensional hyperbolic space $\mathbb{H}^n$ is a Riemannian manifold of constant negative curvature. There exist several isometric models of hyperbolic space, in our work we stick to the Poincar\'e ball model $(\mathbb{D}^n_c, g^{\mathbb{D}})$ with the curvature parameter $c$ (the actual curvature value is then $-c^2$). This model is realized as a pair of an $n$-dimensinal ball $\mathbb{D}^n = \{ x \in \mathbb{R}^n \colon c\|x\|^2 < 1, c \geq 0\} $ equipped with the Riemannian metric $g^{\mathbb{D}} = \lambda_c^2 g^E$, where $ \lambda_c = \frac{2}{1-c\|x\|^2}$ is the \emph{conformal factor} and $g^E = \mathbf{I}_n$ is Euclidean metric tensor. This means, that local distances are scaled by the factor $\lambda_c$ approaching infinity near the boundary of the ball. This gives rise to the `space expansion` property of hyperbolic spaces. While, in the Euclidean spaces, the volume of an object of a diameter $r$ scales polynomially in $r$, in the hyperbolic space, such volumes scale \emph{exponentially} with $r$. Intuitively, this is a continuous analogue of trees: for a tree with a branching factor $k$, we obtain $O(k^d)$ nodes on the level $d$, which in this case serves as a discrete analogue of the radius. This property allows us to efficiently embed hierarchical data even in low dimensions, which is made precise by embedding theorems for trees and complex networks \cite{sarkar2011low}.

% The distance function, intuition on 'space expansion', c \to 1 
Hyperbolic spaces are not vector spaces; to be able to perform operations such as addition, we need to introduce a so-called gyrovector formalism \cite{ungar2008gyrovector}. For a pair $\mathbf{x}, \mathbf{y} \in \mathbb{D}^n_c$, their addition is defined as
\begin{equation}
    \mathbf{x} \oplus_c \mathbf{y} = \frac{(1+2c\langle \mathbf{x}, \mathbf{y} \rangle + c\|\mathbf{y}\|^2) \mathbf{x}+ (1-c\|\mathbf{x}\|^2)\mathbf{y}}{1+2c\langle \mathbf{x}, \mathbf{y} \rangle + c^2 \|\mathbf{x}\|^2 \|\mathbf{y}\|^2}.
\end{equation}

The hyperbolic distance between $\mathbf{x}, \mathbf{y} \in \mathbb{D}^n_c$ is defined in the following manner:
\begin{equation}\label{eq:hdist}
    D_{hyp}(\mathbf{x},\mathbf{y}) = \frac{2}{\sqrt{c}} \mathrm{arctanh}(\sqrt{c}\|-\mathbf{x} \oplus_c \mathbf{y}\|).
\end{equation}

Note that with $c \to 0$ the distance function \eqref{eq:hdist} reduces to Euclidean: $\lim_{c \to 0} D_{hyp}(\mathbf{x},\mathbf{y})=2\|\mathbf{x}-\mathbf{y}\|.$ 

% Exp map, log map for completeness maybe
We also need to define a bijection from Euclidean space to the Poincar\'e model of hyperbolic geometry. This mapping is termed \textit{exponential} while its inverse mapping from hyperbolic space to Euclidean is called \textit{logarithmic}. 

For some fixed base point $\mathbf{x} \in \mathbb{D}^n_c$, the exponential mapping is a function $\exp_\mathbf{x}^c \colon \mathbb{R}^n \to \mathbb{D}_c^n$ defined as:
\begin{equation}\label{eq:exp}
    \exp_\mathbf{x}^c(\mathbf{v}) = \mathbf{x} \oplus_ c \bigg(\tanh \bigg(\sqrt{c} \frac{\lambda_\mathbf{x}^c \|\mathbf{v}\|}{2} \bigg) \frac{\mathbf{v}}{\sqrt{c}\|\mathbf{v}\|}\bigg).
\end{equation}
The base point $\mathbf{x}$ is usually set to $\mathbf{0}$ which makes formulas less cumbersome and empirically has little impact on the obtained results.

% notation $x_i$ - sample - encoder output - project - $z_i$ - embedding
To train our model, we take a sample $\mathbf{x}_i$, pass it through the encoder and project the output to hyperbolic space; the resulted representation in hyperbolic space is denoted as $\mathbf{z}_i$. Since our pairwise cross-entropy loss is based on hyperbolic distances, we do not project $\mathbf{z}_i$ back to Euclidean space and use only the exponential mapping. 

\subsection{Pairwise Cross-Entropy Loss}
\label{sec:xent}

At each iteration, we sample $N$ different categories of images and two samples per category. In this case, the total number of samples (batch size) is $K = 2N$ consisting of $N$ positive pairs.

Additionally to hyperbolic distance, we define the distance with the cosine similarity, implemented with a squared Euclidean distance between normalized vectors:
\begin{equation}
\label{eq:dcos}
D_{cos}(\mathbf{z}_i, \mathbf{z}_j) = \norm{ \frac{\mathbf{z}_i}{\norm{\mathbf{z}_i}_2} - \frac{\mathbf{z}_j}{\norm{\mathbf{z}_j}_2} }^2_2 = 2 - 2 \frac{\langle \mathbf{z}_i, \mathbf{z}_j \rangle}{\norm{\mathbf{z}_i}_2 \cdot \norm{\mathbf{z}_j}_2}
\end{equation}

The loss function for a positive pair $(i, j)$ is defined as

\begin{equation}
\label{eq:xent}
    l_{i, j} = - \log \frac{\exp{(-D(\mathbf{z}_i, \mathbf{z}_j) / \tau)}}{  \sum_{k=1, k \neq i}^K \exp{(-D(\mathbf{z}_i,  \mathbf{z}_k) / \tau) } },
\end{equation}

\noindent
where $D$ is a distance ($D_{hyp}$ or $D_{cos}$) and $\tau$ is a temperature hyperparameter. The total loss is computed for all positive pairs, both $(i, j)$ and $(j, i)$, in a batch.

If the total number of categories is small and a larger batch size is more suitable from the optimization perspective, it is possible to sample more than two samples per category. In this case, we sample $d$ images per each category, $d \geq 2$. We divide the batch $K = dN$ into $d$ subsets with each subset consisting of $N$ samples from different categories. Next, we obtain the loss value for each pair of subsets, as defined \Cref{eq:xent}, summing them up for the final value.

\subsection{$\delta$-hyperbolicity}
\label{sec:delta}
While the curvature value of an underlying manifold for embedding is often neglected, a more efficient way is to estimate it for each dataset specifically. Following the analysis in \cite{khrulkov2020hyperbolic}, we estimate a `measure' of the data hyperbolicity. This evaluation is made through the computation of the so-called Gromov $\delta$.
Its calculation requires first computing $\emph{Gromov product}$ for points $x,y,z \in \mathcal{X}$:
\begin{equation}\label{gromov_product}
(y,z)_x= \frac{1}{2} (d(x,y) +d(x,z) - d(y,z)),
\end{equation}
where $(\mathcal{X},d)$ is an arbitrary metric space. For a set of points, we compute the matrix $M$ of pairwise Gromov products~\eqref{gromov_product}. The $\delta$ value is then defined as the largest entry in the matrix $(M \otimes M) - M$. Here, $\otimes$ denotes the min-max matrix product defined as $(A \otimes B)_{ij} = \max_k \min \{A_{ik}, B_{kj}\}$ \cite{fournier2015computing}.

Being rescaled between $0$ and $1$, the relative $\delta$-hyperbolicity reflects how close to the hyperbolic the hidden structure is: values tending to $0$ show the higher degree of intrinsic data hyperbolicity. The $\delta$ value is related to the optimal radius of the Poincar\'e ball for embeddings through the following expression  $c(X) = (\frac{0.144}{\delta})^2$.
We adopt the procedure described in \cite{khrulkov2020hyperbolic} and evaluate $\delta$ for image embeddings extracted using three encoders: ViT-S, DeiT-S and DINO (described in \Cref{sec:vit}). \Cref{tab:exp_c} highlights the obtained relative $\delta$ values for CUB-200, Cars-196, SOP and In-Shop datasets. %suggesting ... 
\setlength{\tabcolsep}{0.6em}
\begin{table}
  \centering
  \begin{tabular}{lcccc}
    \toprule
    & CUB-200 & Cars-196 & SOP & In-Shop\\
    \midrule
% ResNet-50 & 0.291 & 0.318 & 0.231 & 0.306\\
    ViT-S     & 0.280 & 0.339 & 0.271 & 0.313\\
    DeiT-S    & 0.294 & 0.343 & 0.270 & 0.323\\
    DINO      & 0.315 & 0.327 & 0.301 & 0.318\\
    \bottomrule
  \end{tabular}
  \caption{$\delta$-hyperbolicity values calculated for the embeddings obtained from different encoders. We can see that the $\delta$ values are fairly consistent with respect to different feature extractors. Lower $\delta$ values indicate a higher degree of data hyperbolicity.}
  \label{tab:exp_c}
\end{table}

\subsection{Feature Clipping}
\label{sec:clip}
The paper \cite{guo2021free} empirically shows that a hyperbolic neural network tends to have vanishing gradients since it pushes the embeddings close to the boundary of Poincar\'e ball, making the gradients of Euclidean parameters vanish. To avoid numerical errors when dealing with hyperbolic neural networks, the common approach is to perform clipping by norm on the points in the Poincar\`e ball; the standard norm value is $\frac{1}{\sqrt{c}}(1-10^{-5})$. Instead, the paper \cite{guo2021free} proposes to augment this procedure with an additional technique called \emph{feature clipping}: 
\begin{equation}
    \mathbf{x}^E_C = \min \big\{ 1, \frac{r}{\|\mathbf{x}^E\|} \big\} \cdot \mathbf{x}^E,
\end{equation}
where $\mathbf{x}^E$ lies in the Euclidean space, $\mathbf{x}^E_C$ is its clipped counterpart and $r$ is a new \emph{effective radius} of the Poincar\'e ball. Intuitively, this allows us to push embeddings further away from the boundary and avoid the vanishing gradients problem; in the experiments of \cite{guo2021free} it led to a consistent improvement over baselines.

\subsection{Vision Transformers}
\label{sec:vit}

In our experiments, we use ViT architecture introduced by \cite{vit}. The input image is sliced into patches of size $16 \times 16$ pixels. Each patch is flattened and then linearly projected into an embedding. The resulting vectors are concatenated with position embeddings. Also, this set of vectors includes an additional ``classification'' token. Note that in our case, this token is used to obtain the image embedding, but we do not train a standard classifier as in \cite{vit}. For consistency with previous literature, we name this token \texttt{[class]}. The set of resulting vectors is fed into a standard transformer encoder \cite{attention}. It consists of several layers with multiheaded self-attention (MSA) and MLP blocks, with a \texttt{LayerNorm} before and a residual connection after each block. The output for the transformer encoder for the \texttt{[class]} token is used as the final image representation. For more details, we refer to \cite{vit}.

ViT-S \cite{vit_pretrain} is a smaller version of ViT with 6 heads in MSA (base version uses 12 heads). This architecture is similar to ResNet-50 \cite{resnet} in terms of number of parameters (22M for ViT-S and 23M for ResNet-50) and computational requirements (8.4 FLOPS for ViT-S and 8.3 FLOPS for ResNet-50). This similarity makes it possible to fairly compare with previous works based on ResNet-50 encoder, for this reason, we employ this configuration for our experiments. A more thorough description is available in \cite{vit_pretrain}.

Vision transformers, compared to CNNs, require more training signal. One solution, as proposed in \cite{vit}, is to use a large dataset. ImageNet-21k \cite{imagenet} contains approximately 14M images classified into 21K categories. {\bf ViT-S}, pretrained on ImageNet-21k, is publicly available \cite{vit_pretrain}; we include it in our experiments. Another solution, {\bf DeiT-S} \cite{deit}, is based on the same (ViT-S) architecture and is trained on a smaller ImageNet-1k dataset \cite{imagenet1k} (a subset of ImageNet-21k consisting of about 1.3M training images and 1K categories). An additional training signal is provided by teacher-student distillation, with a CNN-based teacher \cite{deit}.

The third solution used in our experiments, {\bf DINO} \cite{dino}, is based on self-supervised training. In this case, the model ViT-S is trained on the ImageNet-1k dataset \cite{imagenet1k} without labels. The encoder must produce consistent output for different parts of an image, obtained using augmentations (random crop, color jitter, and others). This training scheme is in line with the image retrieval task; in both cases, the encoder is explicitly trained to produce similar output for semantically similar input. However, the goal of these tasks is different: self-supervised learning provides pretrained features, which are then used for other downstream tasks, while for image retrieval resulting features are directly used for the evaluation.

\section{Experiments}
\label{sec:exp}
We follow a widely adopted training and evaluation protocol \cite{proxy_anchor} and compare several versions of our method with current state-of-the-art on four benchmark datasets for category-level retrieval. We include technical details of datasets, our implementation and training details, and finally, present empirical results. There are two types of experiments, first, we compare with the state-of-the-art, and then we investigate the impact of hyperparameters (encoder patch size, manifold curvature, embedding size and batch size).

\subsection{Datasets}
\label{sec:data}
{\bf CUB-200-2011} (CUB) \cite{cub200} includes 11,788 images with 200 categories of bird breeds. The training set corresponds to the first 100 classes with 5,864 images, and the remaining 100 classes with 5,924 images are used for testing. The images are very similar; some breeds can only be distinguished by minor details, making this dataset challenging and, at the same time, informative for the image retrieval task.
{\bf Cars-196} (Cars) \cite{cars196} consists of 16,185 images representing 196 car models. First 98 classes (8,054 images) are used for training and the other 98 classes (8,131 images) are held out 
for testing.
{\bf Stanford Online Product} (SOP) \cite{sop} consists of 120,053 images of 22,634 products downloaded from eBay.com.  We use the standard split: 11,318 classes (59,551 images) for training and remaining 11,316 classes (60,502 images) for testing.
{\bf In-shop Clothes Retrieval} (In-Shop) \cite{inshop} consists of 7,986 categories of clothing items. First 3,997 categories (25,882 images) are for training, the remaining 3,985 categories are used for testing partitioned into a query set (14,218 images) and a gallery set (12,612 images).

\subsection{Implementation Details}
\label{sec:impl}

We use ViT-S \cite{vit_pretrain} as an encoder with three types of pretraining (ViT-S, DeiT-S and DINO), details are presented in \Cref{sec:vit}. The linear projection for patch embeddings as a first basic operation presumably corresponds to low-level feature extraction, so we freeze it during fine-tuning. The encoder outputs a representation of dimensionality $384$, which is further plugged into a head linearly projecting the features to the space of dimension $128$. We initialize the biases of the head with constant $0$ and weights with a (semi) orthogonal matrix \cite{ortho_init}. We include two versions of the head: with a projection to a hyperbolic space (``Hyp-'') and with projection to a unit hypersphere (``Sph-''). In the first case, we use curvature parameter $c = 0.1$ (in \Cref{sec:ablations} we investigate how it affects the method's performance), temperature $\tau = 0.2$ and clipping radius (defined in \Cref{sec:clip}) $r = 2.3$. For spherical embeddings, we use temperature $\tau = 0.1$.

To evaluate the model performance, for the encoder, we compute the Recall@K metric for the output with distance $D_{cos}$ (\cref{eq:dcos}); for the head, we use $D_{cos}$ for ``Sph-'' version and hyperbolic distance $D_{hyp}$ (\cref{eq:hdist}) for ``Hyp-'' version. We resize the test images to $224$ ($256$ for CUB) on the smaller side and take one $224 \times 224$ center crop. Note that some methods use images of higher resolution for training and evaluations, \eg, ProxyNCA++ \cite{PNCAPP} use $256 \times 256$ crops indicating that smaller $227 \times 227$ crops degrade the performance by $4.3\%$ on CUB. However, $224\times 224$ is the default size for encoders considered in our work; moreover, some recent methods, such as IRT\textsubscript{R} \cite{IRT}, use this size for experiments.

We use the AdamW optimizer \cite{adamw} with a learning rate value $1 \times 10^{-5}$ for DINO and $3 \times 10^{-5}$ for ViT-S and DeiT-S. The weight decay value is $0.01$, and the batch size equals $900$. The number of optimizer steps depends on the dataset: $200$ for CUB, $600$ for Cars, $25000$ for SOP, $2200$ for In-Shop. The gradient is clipped by norm $3$ for a greater stability. We apply commonly used data augmentations: random crop resizing the image to $224 \times 224$ using bicubic interpolation combined with a random horizontal flip. We train with Automatic Mixed Precision in \texttt{O2} mode \footnote{\url{https://github.com/NVIDIA/apex}}. All experiments are performed on one NVIDIA A100 GPU.

\subsection{Results}
\label{sec:results}

\setlength{\tabcolsep}{0.47em}
\begin{table*}
  \centering
  \begin{tabular}{l|cccc|cccc|cccc|cccc}
    \toprule
    \multirow{2}{*}{Method} &
    \multicolumn{4}{c|}{CUB-200-2011 (K)} &
    \multicolumn{4}{c|}{Cars-196 (K)} &
    \multicolumn{4}{c|}{SOP (K)} &
    \multicolumn{4}{c}{In-Shop (K)} \\
    &
    1 & 2 & 4 & 8 &
    1 & 2 & 4 & 8 &
    1 & 10 & 100 & 1000 &
    1 & 10 & 20 & 30 \\
    \midrule
    % ProxyNCA \cite{ProxyNCA} &
    % 49.2 & 61.9 & 67.9 & 72.4 &
    % 73.2 & 82.4 & 86.4 & 88.7 &
    % 73.7 & - & - & - &
    % - & - & - & - \\
    Margin \cite{Margin} &
    63.9 & 75.3 & 84.4 & 90.6 &
    79.6 & 86.5 & 91.9 & 95.1 &
    72.7 & 86.2 & 93.8 & 98.0 &
    - & - & - & - \\
    FastAP \cite{FastAP} &
    - & - & - & - &
    - & - & - & - &
    73.8 & 88.0 & 94.9 & 98.3 &
    - & - & - & - \\
    NSoftmax \cite{NSoftmax} &
    56.5 & 69.6 & 79.9 & 87.6 &
    81.6 & 88.7 & 93.4 & 96.3 &
    75.2 & 88.7 & 95.2 & - &
    86.6 & 96.8 & 97.8 & 98.3 \\
    MIC \cite{MIC} &
    66.1 & 76.8 & 85.6 & - &
    82.6 & 89.1 & 93.2 & - &
    77.2 & 89.4 & 94.6 & - &
    88.2 & 97.0 & - & 98.0 \\
    XBM \cite{XBM} &
    - & - & - & - &
    - & - & - & - &
    80.6 & 91.6 & 96.2 & 98.7 &
    91.3 & 97.8 & 98.4 & 98.7 \\
    IRT\textsubscript{R} \cite{IRT} &
    72.6 & 81.9 & 88.7 & 92.8 &
    - & - & - & - &
    83.4 & 93.0 & 97.0 & 99.0 &
    91.1 & 98.1 & 98.6 & 99.0 \\
    \midrule
    Sph-DeiT &
    73.3 & 82.4 & 88.7 & 93.0 &
    77.3 & 85.4 & 91.1 & 94.4 &
    82.5 & 93.1 & 97.3 & 99.2 &
    89.3 & 97.0 & 97.9 & 98.4 \\
    Sph-DINO &
    76.0 & 84.7 & 90.3 & 94.1 &
    81.9 & 88.7 & 92.8 & 95.8 &
    82.0 & 92.3 & 96.9 & 99.1 &
    90.4 & 97.3 & 98.1 & 98.5 \\
    Sph-ViT \textsuperscript{$\mathsection$} &
    83.2 & 89.7 & 93.6 & 95.8 &
    78.5 & 86.0 & 90.9 & 94.3 &
    82.5 & 92.9 & 97.4 & 99.3 &
    90.8 & 97.8 & 98.5 & 98.8 \\
    Hyp-DeiT &
    74.7 & 84.5 & 90.1 & 94.1 &
    82.1 & 89.1 & 93.4 & 96.3 &
    83.0 & 93.4 & 97.5 & 99.2 &
    90.9 & 97.9 & 98.6 & 98.9 \\
    Hyp-DINO &
    78.3 & 86.0 & 91.2 & 94.7 &
    {\bf 86.0} & {\bf 91.9} & {\bf 95.2} & {\bf 97.2} &
    84.6 & 94.1 & 97.7 & 99.3 &
    92.6 & {\bf 98.4} & {\bf 99.0} & {\bf 99.2} \\
    Hyp-ViT \textsuperscript{$\mathsection$} &
    {\bf 84.0} & {\bf 90.2} & {\bf 94.2} & {\bf 96.4} &
    82.7 & 89.7 & 93.9 & 96.2 &
    {\bf 85.5} & {\bf 94.9} & {\bf 98.1} & {\bf 99.4} &
    {\bf 92.7} & {\bf 98.4} & 98.9 & 99.1 \\
    \bottomrule
  \end{tabular}
  \caption{Recall@K metric for four datasets for 128-dimensional embeddings. The 6 versions of our method are listed in the bottom section, evaluated for head embeddings. \mbox{``Sph-''} are versions with hypersphere embeddings optimised using $D_{cos}$ (\cref{eq:dcos}), ``Hyp-'' are versions with hyperbolic embeddings optimised using $D_{hyp}$ (\cref{eq:hdist}). ``DeiT'', ``DINO'' and ``ViT'' indicate type of pretraining for the vision transformer encoder. Margin, FastAP, MIC, XBM, NSoftmax are based on ResNet-50 \cite{resnet} encoder, IRT\textsubscript{R} is based on DeiT \cite{deit}.\\
  \textsuperscript{$\mathsection$} pretrained on the larger ImageNet-21k \cite{imagenet}.
  }
  \label{tab:exp_128}
\end{table*}

\setlength{\tabcolsep}{0.35em}
\begin{table*}
  \centering
  \begin{tabular}{l|c|cccc|cccc|cccc|cccc}
    \toprule
    \multirow{2}{*}{Method} &
    \multirow{2}{*}{Dim} &
    \multicolumn{4}{c|}{CUB-200-2011 (K)} &
    \multicolumn{4}{c|}{Cars-196 (K)} &
    \multicolumn{4}{c|}{SOP (K)} &
    \multicolumn{4}{c}{In-Shop (K)} \\
    & &
    1 & 2 & 4 & 8 &
    1 & 2 & 4 & 8 &
    1 & 10 & 100 & 1000 &
    1 & 10 & 20 & 30 \\
    \midrule
    A-BIER \cite{A_BIER} & 512 &
    57.5 & 68.7 & 78.3 & 86.2 &
    82.0 & 89.0 & 93.2 & 96.1 &
    74.2 & 86.9 & 94.0 & 97.8 &
    83.1 & 95.1 & 96.9 & 97.5 \\
    ABE \cite{ABE} & 512 &
    60.6 & 71.5 & 79.8 & 87.4 &
    85.2 & 90.5 & 94.0 & 96.1 &
    76.3 & 88.4 & 94.8 & 98.2 &
    87.3 & 96.7 & 97.9 & 98.2 \\
    SM \cite{SM} & 512 &
    56.0 & 68.3 & 78.2 & 86.3 &
    83.4 & 89.9 & 93.9 & 96.5 &
    75.3 & 87.5 & 93.7 & 97.4 &
    90.7 & 97.8 & 98.5 & 98.8 \\
    XBM \cite{XBM} & 512 &
    65.8 & 75.9 & 84.0 & 89.9 &
    82.0 & 88.7 & 93.1 & 96.1  &
    79.5 & 90.8 & 96.1 & 98.7 &
    89.9 & 97.6 & 98.4 & 98.6 \\
    HTL \cite{HTL} & 512 &
    57.1 & 68.8 & 78.7 & 86.5 &
    81.4 & 88.0 & 92.7 & 95.7 &
    74.8 & 88.3 & 94.8 & 98.4 &
    80.9 & 94.3 & 95.8 & 97.2 \\
    MS \cite{MS} & 512 &
    65.7 & 77.0 & 86.3 & 91.2 &
    84.1 & 90.4 & 94.0 & 96.5 &
    78.2 & 90.5 & 96.0 & 98.7 &
    89.7 & 97.9 & 98.5 & 98.8 \\
    SoftTriple \cite{softtriple} & 512 &
    65.4 & 76.4 & 84.5 & 90.4 &
    84.5 & 90.7 & 94.5 & 96.9 &
    78.6 & 86.6 & 91.8 & 95.4 &
    - & - & - & - \\
    HORDE \cite{HORDE} & 512 &
    66.8 & 77.4 & 85.1 & 91.0 &
    86.2 & 91.9 & 95.1 & 97.2 &
    80.1 & 91.3 & 96.2 & 98.7 &
    90.4 & 97.8 & 98.4 & 98.7 \\
    Proxy-Anchor \cite{proxy_anchor} & 512 &
    68.4 & 79.2 & 86.8 & 91.6 &
    86.1 & 91.7 & 95.0 & 97.3 &
    79.1 & 90.8 & 96.2 & 98.7 &
    91.5 & 98.1 & 98.8 & {\bf 99.1} \\
    NSoftmax \cite{NSoftmax} & 512 &
    61.3 & 73.9 & 83.5 & 90.0 &
    84.2 & 90.4 & 94.4 & 96.9 &
    78.2 & 90.6 & 96.2 & - & 
    86.6 & 97.5 & 98.4 & 98.8 \\
    ProxyNCA++ \cite{PNCAPP} & 512 &
    69.0 & 79.8 & 87.3 & 92.7 &
    86.5 & 92.5 & 95.7 & 97.7 &
    80.7 & 92.0 & 96.7 & 98.9 &
    90.4 & 98.1 & 98.8 & 99.0 \\
    % NSoftmax \cite{NSoftmax} & 2048 &
    % 65.3 & 76.7 & 85.4 & 91.8 &
    % 89.3 & 94.1 & 96.4 & 98.0 &
    % 79.5 & 91.5 & 96.7 & - &
    % 89.4 & 97.8 & 98.7 & 99.0 \\
    % ProxyNCA++ \cite{PNCAPP} & 2048 &
    % 72.2 & 82.0 & 89.2 & 93.5 &
    % 90.1 & 94.5 & 97.0 & 98.4 &
    % 81.4 & 92.4 & 96.9 & 99.0 &
    % 90.9 & 98.2 & 98.9 & 99.1 \\
    IRT\textsubscript{R} \cite{IRT} & 384 &
    76.6 & 85.0 & 91.1 & 94.3 &
    - & - & - & - &
    84.2 & 93.7 & 97.3 & 99.1 &
    91.9 & 98.1 & 98.7 & 98.9 \\
    \midrule
    ResNet-50 \cite{resnet} \textsuperscript{\dag} & 2048 &
    41.2 & 53.8 & 66.3 & 77.5 &
    41.4 & 53.6 & 66.1 & 76.6 &
    50.6 & 66.7 & 80.7 & 93.0 &
    25.8 & 49.1 & 56.4 & 60.5 \\
    DeiT-S \cite{deit} \textsuperscript{\dag} & 384 &
    70.6 & 81.3 & 88.7 & 93.5 &
    52.8 & 65.1 & 76.2 & 85.3 &
    58.3 & 73.9 & 85.9 & 95.4 &
    37.9 & 64.7 & 72.1 & 75.9 \\
    DINO \cite{dino} \textsuperscript{\dag} & 384 &
    70.8 & 81.1 & 88.8 & 93.5 &
    42.9 & 53.9 & 64.2 & 74.4 &
    63.4 & 78.1 & 88.3 & 96.0 &
    46.1 & 71.1 & 77.5 & 81.1 \\
    ViT-S \cite{vit_pretrain} \textsuperscript{\dag} \textsuperscript{$\mathsection$} & 384 &
    83.1 & 90.4 & 94.4 & 96.5 &
    47.8 & 60.2 & 72.2 & 82.6 &
    62.1 & 77.7 & 89.0 & 96.8 &
    43.2 & 70.2 & 76.7 & 80.5 \\
    \midrule
    Sph-DeiT & 384 &
    76.2 & 84.5 & 90.2 & 94.3 &
    81.7 & 88.6 & 93.4 & 96.2 &
    82.5 & 92.9 & 97.2 & 99.1 &
    89.6 & 97.2 & 98.0 & 98.4 \\
    Sph-DINO & 384 &
    78.7 & 86.7 & 91.4 & 94.9 &
    86.6 & 91.8 & 95.2 & 97.4 &
    82.2 & 92.1 & 96.8 & 98.9 &
    90.1 & 97.1 & 98.0 & 98.4 \\
    Sph-ViT \textsuperscript{$\mathsection$} & 384 &
    85.1 & 90.7 & 94.3 & 96.4 &
    81.7 & 89.0 & 93.0 & 95.8 &
    82.1 & 92.5 & 97.1 & 99.1 &
    90.4 & 97.4 & 98.2 & 98.6 \\
    Hyp-DeiT & 384 &
    77.8 & 86.6 & 91.9 & 95.1 &
    86.4 & 92.2 & 95.5 & 97.5 &
    83.3 & 93.5 & 97.4 & 99.1 &
    90.5 & 97.8 & 98.5 & 98.9 \\
    Hyp-DINO & 384 &
    80.9 & 87.6 & 92.4 & 95.6 &
    {\bf 89.2} & {\bf 94.1} & {\bf 96.7} & {\bf 98.1} &
    85.1 & 94.4 & 97.8 & 99.3 &
    92.4 & {\bf 98.4} & {\bf 98.9} & {\bf 99.1} \\
    Hyp-ViT \textsuperscript{$\mathsection$} & 384 &
    {\bf 85.6} & {\bf 91.4} & {\bf 94.8} & {\bf 96.7} &
    86.5 & 92.1 & 95.3 & 97.3 &
    {\bf 85.9} & {\bf 94.9} & {\bf 98.1} & {\bf 99.5} &
    {\bf 92.5} & 98.3 & 98.8 & {\bf 99.1} \\
    \bottomrule
  \end{tabular}
  \caption{
    Recall@K metric for four datasets, ``Dim'' column shows the dimensionality of embeddings. The 6 versions of our method are listed in the bottom section, evaluated for encoder embeddings, titles are described in \cref{tab:exp_128}. Encoders by method: A-BIER, ABE, SM: GoogleNet \cite{GoogleNet}; XBM, HTL, MS, SoftTriple, HORDE, Proxy-Anchor: Inception with batch normalization \cite{batch_norm}; NSoftmax, ProxyNCA++: ResNet-50 \cite{resnet}; IRT\textsubscript{R}: DeiT \cite{deit}.
    \textsuperscript{\dag} pretrained encoders without training on the target dataset. \textsuperscript{$\mathsection$} pretrained on the larger ImageNet-21k \cite{imagenet}.
    }
  \label{tab:exp_big}
\end{table*}

\Cref{tab:exp_128} highlights the experimental results for the 128-dimensional head embedding and the results for 384-dimensional encoder embedding are shown in \Cref{tab:exp_big}. We include evaluation of the pretrained encoders without training on the target dataset in \Cref{tab:exp_big} for reference. On the CUB dataset, we can observe the solid performance of methods with ViT encoder; the gap between the second-best method IRT\textsubscript{R} and Hyp-ViT is $9\%$. However, the main improvement comes from the dataset used for pretraining (ImageNet-21k), since Hyp-DINO and Hyp-DeiT demonstrate a smaller improvement, while baseline ViT-S without finetuning shows strong performance. We hypothesize that this is due to the presence of several bird classes in the ImageNet-21k dataset encouraging the encoder to separate them during the pretraining phase.

For the SOP and In-Shop datasets, the difference between Hyp-ViT and Hyp-DINO is minor, while, for Cars-196, Hyp-DINO outperforms Hyp-ViT with a significant margin. These results confirm that both pretraining schemes are suitable for the considered task.
%It is worth noting that Hyp-DINO was pretrained with self-supervised learning method \cite{dino} without labels, opening applications of our method in many domains, where labels are difficult to acquire. One justification of the best performance of the DINO encoder on the Cars-196 dataset is that the encoder is explicitly trained on the images with color jitter augmentations enforcing the color invariance, probably making it easier to correctly classify models of cars independently of colors. 
The versions with DeiT perform worse compared to ViT- and DINO-based encoders while outperforming CNN-based models. This observation confirms the significance of vision transformers in our architecture.
The experimental results suggest that hyperbolic space embeddings consistently improve the performance compared to spherical versions. Hyperbolic space seems to be beneficial for the embeddings, and the distance in hyperbolic space suits well for the pairwise cross-entropy loss function. At the same time, our sphere-based versions perform well compared to other methods with CNN encoders.

\begin{figure}
\begin{center}
\includegraphics[width=1\linewidth]{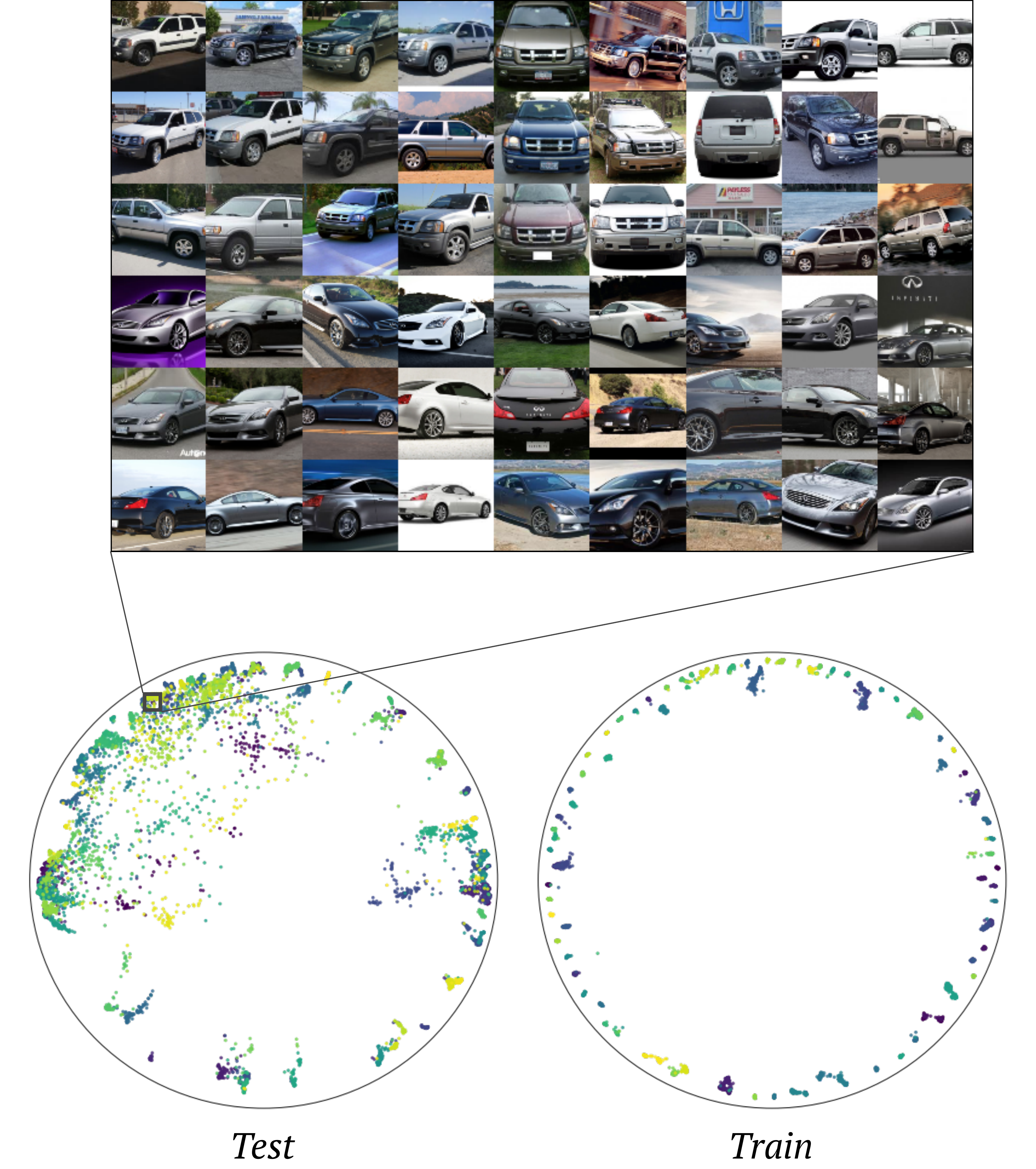}
\vspace{-2mm}
\caption{Hyp-DINO embeddings for Cars-196 dataset (training and evaluation sets) on the Poincar\'e disk. Each point inside the disk corresponds to a sample, different colors indicate different classes. Images of cars are plotted preserving neighborhood relations of samples.}
\label{fig:cars}
\end{center}
\end{figure}

\Cref{fig:cars} illustrates how learned embeddings are arranged on the Poincar\'e disk. We use UMAP \cite{UMAP} method with the ``hyperboloid'' distance metric to reduce the dimensionality to 2D for visualization. For the training part, we can see that samples are clustered according to labels, and each cluster is pushed closer to the border of the disk, indicating that the encoder separates classes well. However, for the testing part, the structure is more complex. We observe that some of the samples tend to move towards the center and intermix, while others stay in clusters, showing possible hierarchical relationships. We can see that car images are grouped by several properties: pose, color, shape, etc.

\subsection{Impact of Hyperparameters}
\label{sec:ablations}

In this section, we investigate the impact of the values of the hyperparameters on the model performance.

\setlength{\tabcolsep}{.5em}
\begin{table}
  \centering
  \begin{tabular}{l|c|c c c c}
    \toprule
    \multirow{2}{*}{Method} &
    \multirow{2}{*}{Dim} &
    \multicolumn{4}{c}{Recall@K}\\
    & & 1 & 2 & 4 & 8\\
    \midrule
    NSoftmax \cite{NSoftmax} & 2048 & 89.3 & 94.1 & 96.4 & 98.0 \\
    ProxyNCA++ \cite{PNCAPP} & 2048 & 90.1 & 94.5 & 97.0 & 98.4 \\
    \midrule
    Hyp-DINO $16 \times 16$ & 128 & 86.0 & 91.9 & 95.2 & 97.2 \\
    Hyp-DINO $8 \times 8$ & 128 & 90.4 & 94.7 & 97.0 & 98.2 \\
    Hyp-DINO $16 \times 16$ & 384 & 89.2 & 94.1 & 96.7 & 98.1 \\
    Hyp-DINO $8 \times 8$ & 384 & 92.8 & 96.2 & 97.8 & 98.8 \\
    \bottomrule
  \end{tabular}
  \caption{First two rows represent current best overall result for Cars-196 dataset with ResNet-50 encoder. Our method \mbox{(Hyp-DINO)} is presented with $8 \times 8$ and $16 \times 16$ patch sizes.}
  \label{tab:exp_patch8}
  \vspace{-0.5cm}
\end{table}

{\bf Encoder patch size}. ViT architecture does not process each pixel independently; for computational feasibility, the input image is sliced into patches projected into the initial embeddings. The default size of the patch is $16 \times 16$, although considering other values is also possible. The experiments in \cite{dino} have demonstrated a significant performance gain from smaller $8 \times 8$ patches for self-supervised learning. In this case, the number of parameters of the encoder does not change; however, it requires processing $4\times$ more embeddings, which allows the encoder to learn more complex dependencies between patches. We add an experiment with this setup in \Cref{tab:exp_patch8} demonstrating a substantial performance improvement ($+4.4\%$) compared to the default configuration. In this case, we use the same training procedure, described in \Cref{sec:impl}, with the batch size equal to $120$.

\setlength{\tabcolsep}{1em}
\begin{table}
  \centering
  \begin{tabular}{l|cccc}
    \toprule
    Parameter & Encoder(384) & Head \\
    \midrule
    Default & 92.4 & 92.6 \\
    \midrule
    $c = 0.01$ & 92.3 & 92.6 \\
    $c = 0.05$ & 92.4 & 92.6 \\
    $c = 0.3$ & 92.3 & 92.0 \\
    $c = 0.5$ & 91.8 & 91.0 \\
    $c = 1.0$ & 90.0 & 89.2 \\
    \midrule
    Head dim. $16$ & 88.6 & 83.3 \\
    Head dim. $32$ & 90.2 & 89.6 \\
    Head dim. $64$ & 91.6 & 91.7 \\
    \midrule
    Batch size $200$  & 92.0 & 91.9 \\
    Batch size $400$  & 92.5 & 92.5 \\
    Batch size $1600$ & 92.4 & 92.6 \\
    \bottomrule
  \end{tabular}
  \caption{Recall@1 metric for various hyperparameters for Hyp-ViT configuration on In-Shop dataset. Default configuration is $c = 0.1$, head dimensionality $128$, batch size $900$. }
  \label{tab:exp_param}
  \vspace{-0.6cm}
\end{table}

{\bf Manifold curvature}.
\Cref{tab:exp_param} shows the model performance depending on the curvature value $c$. We observe that the method is robust in the range $(0.01, 0.3)$ while larger values lead to degradation. Notably, the accuracy of the head degrades faster since the hyperbolic distance is also used in the evaluation and the imprecision in this parameter immediately affects the output. The radius of the ball is inversely proportional to the $c$ value. Intuitively, if the $c$ value tends to $0$, the radius tends to infinity, making the ball as flat as the Euclidean space; in contrast, larger $c$ values correspond to a steeper configuration.
Note that according to $\delta$ values (\Cref{tab:exp_c}), the estimated value of c is close to $0.2$, depending on the dataset and encoder. However, smaller values tend to provide better stability; we believe this is due to an optimisation process that can be improved for the hyperbolic space. For this reason, we adjusted the default value towards a smaller $0.1$ (\Cref{sec:impl}).

{\bf Embedding size and batch size}.
As expected, lower output dimensionality leads to lower recall values. However, taking into account a high data variability (3,985 categories in the test set), the experimental results suggest that the method has a reasonable representation power even in the case of lower dimensions.

The batch size directly influences the number of negative examples during the training phase; thus, intuitively, larger values have to be more profitable for the model performance. However, as the experiments show (\Cref{tab:exp_param}), the method is robust for batch size $\geq 400$, having a minor accuracy degradation for batch size equal to $200$. Therefore, for considered datasets, the method does not require distributed training with a large number of GPUs \cite{simclr} or specific solutions with a momentum network \cite{moco}.

\section{Related Work}
\label{sec:related}
{\bf Hyperbolic embeddings.} Learning embeddings in hyperbolic spaces have emerged since this approach was proposed for NLP tasks \cite{nickel2017poincare,nickel2018learning}. Shortly after that, hyperbolic neural networks were presented as a generalization of standard Euclidean operations allowing to learn the data representations directly in hyperbolic spaces \cite{ganea2019hyperbolic}. The authors generalized standard linear layers to hyperbolic counterparts, defined multinomial logistic regression and recurrent neural networks. Several studies showed the benefits of hyperbolic embeddings of visual data when applied to few-shot \cite{khrulkov2020hyperbolic,gao2021curvature,fang2021kernel} and zero-shot learning \cite{liu2020hyperbolic,fang2021kernel}. In \cite{khrulkov2020hyperbolic}, the authors proposed a hybrid architecture with the main bulk of the layers operating in Euclidean space and only final layers operating in hyperbolic space. In \cite{fang2021kernel}, the authors instead focus on kernelization widely used in Euclidean space and generalize them for hyperbolic representations. The paper \cite{liu2020hyperbolic} proposes a method directly incorporating the hierarchical relations for hyperbolic embeddings in application to zero-shot learning.

{\bf Vision transformers in metric learning.}
The paper \cite{IRT} has recently demonstrated beneficial properties of vision transformers for category-level and object retrieval tasks. The proposed IRT\textsubscript{R} employs the architecture and pretraining scheme of DeiT \cite{deit}. The method is trained using the contrastive loss with cross-batch memory \cite{XBM} with momentum encoder \cite{moco} in several experiments. Moreover, the method requires a sophisticated entropy regularization to spread the embeddings more uniformly on a hypersphere. The study performed in \cite{uniformity} has shown that pairwise cross-entropy loss, considered in our work, already possesses this property. 
Asymptotically, this loss can be decoupled into two components: one optimizes the alignment of positive pairs, while the second preserves overall uniformity. Furthermore, the exponential expansion of the volume in the hyperbolic space can facilitate uniform feature alignment.

Self-supervised learning is similar in spirit to metric learning: in both cases, the encoder is trained to produce similar representations for semantically similar images. Consequently, there are various relevant approaches in these domains. DINO \cite{dino} is a recently proposed method, where vision transformer is trained in the self-supervised learning setting. This method shows a high k-NN classification accuracy for obtained representations while also performing well in the image retrieval task. These results suggest that both vision transformers and self-supervised pretraining are advantageous for metric learning and our experiments confirm this.

{\bf Metric learning loss functions.}
A contrastive loss \cite{contrastive_2006} and its popular variation triplet margin loss \cite{triplet} are classic metric learning loss functions. In the first case, the distance between positive pairs is optimized to be lower than some predefined threshold and larger for the negative pairs. The triplet margin loss penalizes the cases where negative examples are closer to each other than positives plus a margin $m$. Another variation of the contrastive loss is the lifted structure loss \cite{sop} with LogSumExp applied to all negative pair distances. Similarly, NCA loss \cite{NCA} minimizes the distance between positives with respect to a set of negatives using exponential weighting. In essence, pairwise cross-entropy loss (\Cref{eq:xent}) equals NCA when all batch samples are used as negatives.

A cross-entropy loss in the form of pairwise-distance loss for the metric learning was introduced by \cite{NPair} as N-pair loss. In addition, the paper \cite{boudiaf2020unifying} established a connection between the standard cross-entropy loss for classification and metric learning losses, proposing their own version of the pairwise cross-entropy loss. Recently, this loss function has seen overwhelming success in the self-supervised learning field \cite{CPC,moco,simclr}. Popular implementations are InfoNCE \cite{CPC} and NT-Xent \cite{simclr}. However, most of these works only consider Euclidean distances between embeddings (generally $L_{2}$-normalized). Our method extends this widely used loss function to the hyperbolic space.

\section{Conclusion}
\label{sec:conclusion}

In this paper, we have combined several improvements for the metric learning task: pairwise cross-entropy loss with the hyperbolic distance function, vision transformers with several pretraining schemes. We empirically verified that each proposed component is crucial for the best performance.
%We have evaluated six versions of our method on four datasets, achieving the new state-of-the-art results. Our model has the same number of parameters and computational requirements as the commonly used architecture ResNet-50 \cite{resnet}. Additionally, we have provided a more computationally demanding version improving the performance even more.
In deep learning tasks, it is often tricky to empirically distinguish the source of improvement between methodological contribution and a technical solution \cite{reality_check}. To address this obstacle, we have presented several versions, comparing elements of our method in the equal setup. The {\em Hyp-DINO} version is of particular interest: it demonstrates that self-supervised learning and metric learning complement each other perfectly, resulting in a powerful metric with minimal supervision.

{\bf Limitations.}
In this work, we have considered only a vision domain with a category-level retrieval task. However, the proposed method is not limited to such applications. Hyperbolic embeddings \cite{nickel2017poincare} and transformers \cite{attention} were initially proposed for the natural language processing. The recently proposed method \cite{openai_clip} shows that combining visual and language domains gives rise to learning universal representation. At the same time, papers \cite{perceiver,lu2021pretrained} demonstrate that one transformer architecture is suitable for many domains at once. Hence, combining multiple domains with a common semantic metric can be an interesting development of this work.

{\bf Broader Impact.}
Common applications in the field of metric learning are face recognition \cite{sphereface,facenet} and person re-identification \cite{chen2017triplet,xiao2017joint}. Such systems can improve people's safety and quality of life, but there are also notorious cases based on gathering personal information. Another possible risk is learned social biases. While this topic is commonly studied in the NLP field, the situation with vision transformers is more subtle and far less explored.
% Common applications in the field of metric learning are face recognition \cite{sphereface,facenet} and person re-identification \cite{chen2017triplet,xiao2017joint}. Such systems can improve people's safety and quality of life, \eg accessing buildings using only facial recognition. However, there are also notorious cases: such systems allow gathering personal information and lead to excessive surveillance. Looking at future prospects, such methods can help search for similarities in a complex information space, moving towards more general and advanced information processing systems.

{\bf Acknowledgments. }
This work was supported by the EU H2020 AI4Media Project under Grant 951911, by the Analytical center under the RF Government (subsidy agreement 000000D730321P5Q0002, Grant No. 70-2021-00145 02.11.2021). We thank Google Cloud Platform (GCP) for providing computing support.

{\small
\bibliographystyle{ieee_fullname}
\bibliography{_main}
}

\end{document}

\title{Hyperbolic Vision Transformers: Combining Improvements in Metric Learning \\
Supplementary}

\maketitle
\thispagestyle{empty}
\appendix

%%%%%%%%% BODY TEXT - ENTER YOUR RESPONSE BELOW
\section{Things we tried but they did not work}

Our initial experiments were focused on {\bf self-supervised learning} (SSL) in hyperbolic space. However, we noticed that the head output, which is usually ignored in SSL formulation, shows high performance, thus we switched to a more suitable metric learning formulation. During preliminary experiments, we tried our method with the {\bf ResNet-50} \cite{resnet} backbone. In this case, the hyperbolic version also outperforms the sphere-based version. However, we do not publish and compare CNN-based architecture with other methods, because the transformer backbone performs clearly better without drawbacks, and we focus on it. We had a modification of our method with the {\bf MoCo} \cite{moco} loss. However, it performed similarly to plain cross-entropy loss, so we decided not to include it in the final version. Also, we tried our method with {\bf ProxyNCA} \cite{ProxyNCA} loss, which performed worse.

\section{Datasets visualization}
\begin{figure}
\begin{center}
\includegraphics[width=1\linewidth]{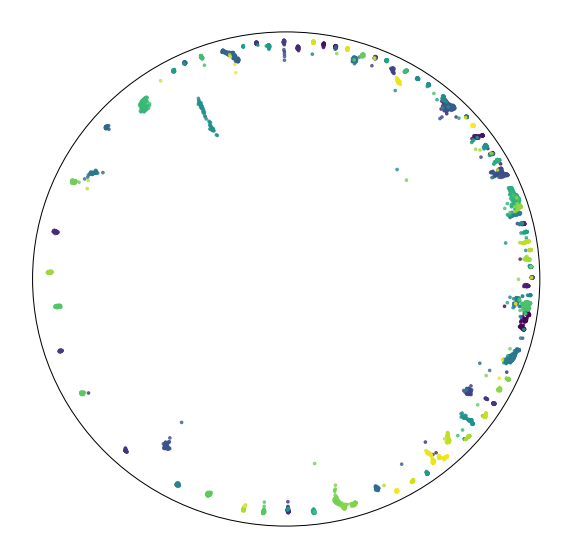}
\vspace{-2mm}
\caption{CUB-200-2011 train set}
\label{fig:cub_train}
\end{center}
\end{figure}

\begin{figure}
\begin{center}
\includegraphics[width=1\linewidth]{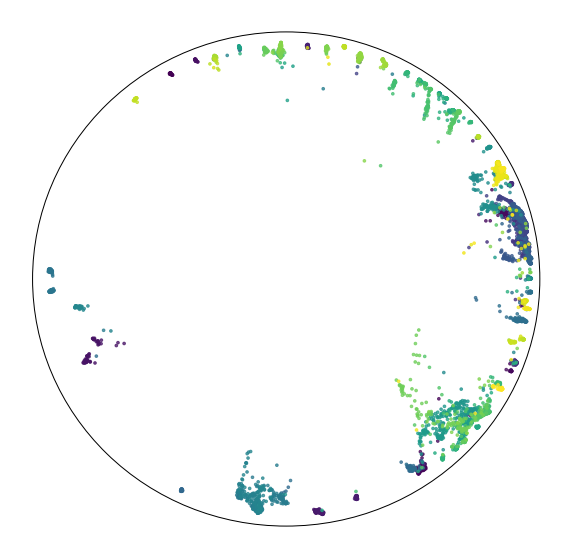}
\vspace{-2mm}
\caption{CUB-200-2011 test set}
\label{fig:cub_eval}
\end{center}
\end{figure}

\begin{figure}
\begin{center}
\includegraphics[width=1\linewidth]{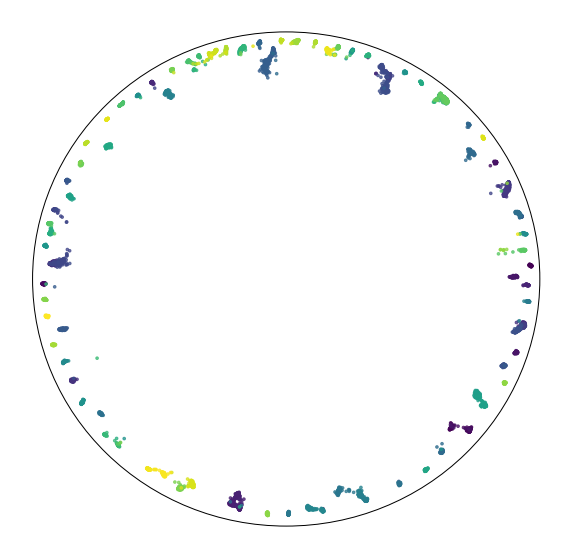}
\vspace{-2mm}
\caption{Cars-196 train set}
\label{fig:cars_train}
\end{center}
\end{figure}

\begin{figure}
\begin{center}
\includegraphics[width=1\linewidth]{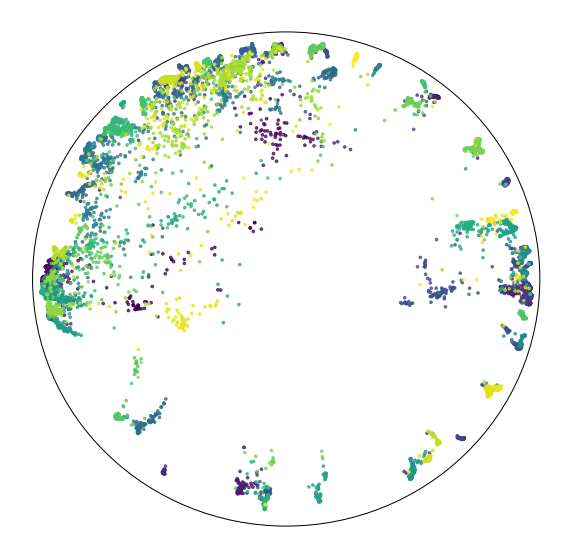}
\vspace{-2mm}
\caption{Cars-196 test set}
\label{fig:cars_eval}
\end{center}
\end{figure}

\Cref{fig:cub_train,fig:cub_eval,fig:cars_train,fig:cars_eval} illustrate how learned embeddings are arranged on the Poincar\'e disk. We use UMAP \cite{UMAP} method with the ``hyperboloid'' distance metric to reduce the dimensionality to 2D for visualization. Embeddings are obtained with Hyp-DINO configuration for CUB-200-2011 and Cars-196 datasets. Each point inside the disk corresponds to a sample, different colors indicate different classes.

\begin{figure*}
\begin{center}
\includegraphics[width=.8\linewidth]{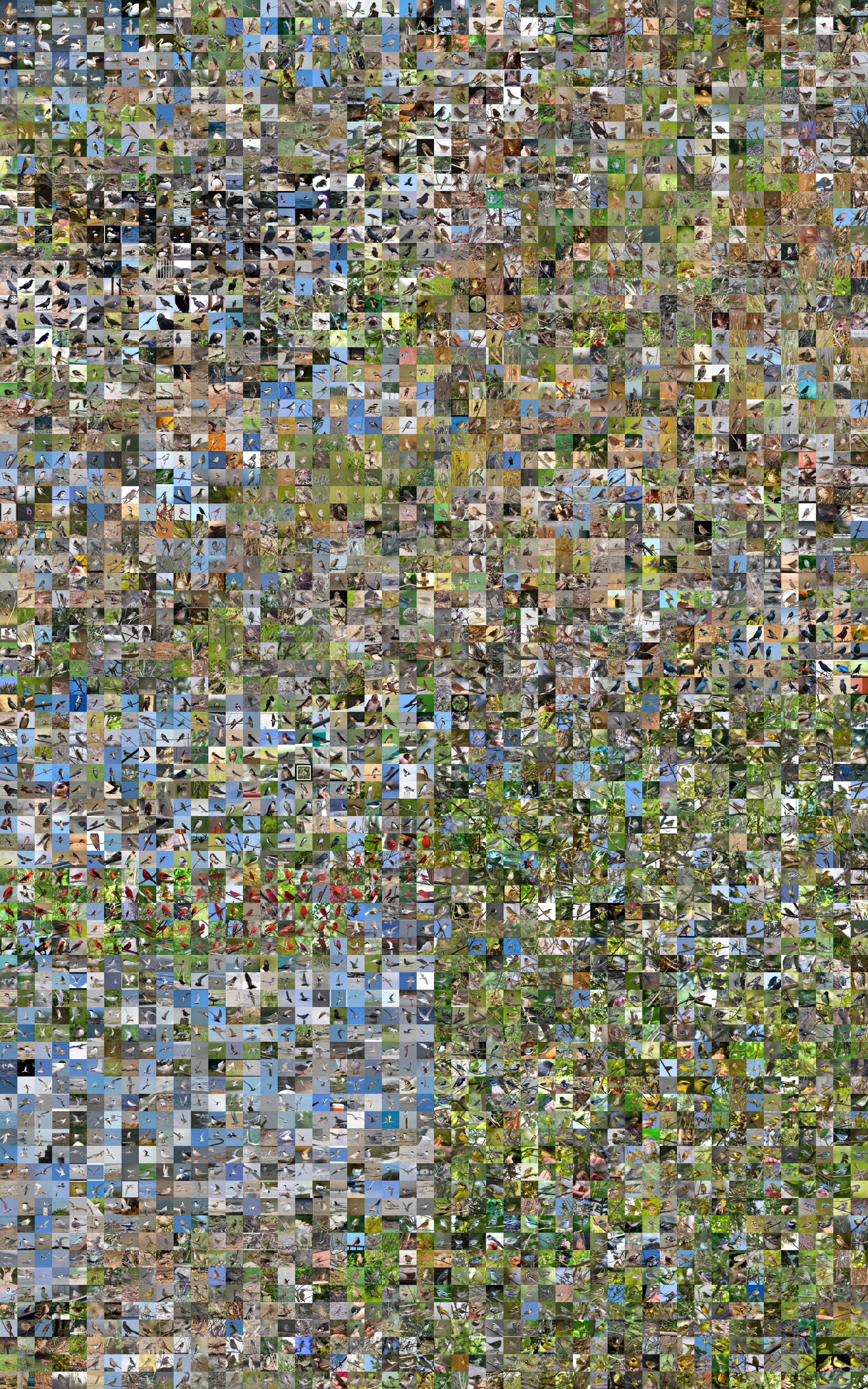}
\vspace{-2mm}
\caption{CUB-200-2011 test subset (4000 images)}
\label{fig:cub_img}
\end{center}
\end{figure*}

\begin{figure*}
\begin{center}
\includegraphics[width=.8\linewidth]{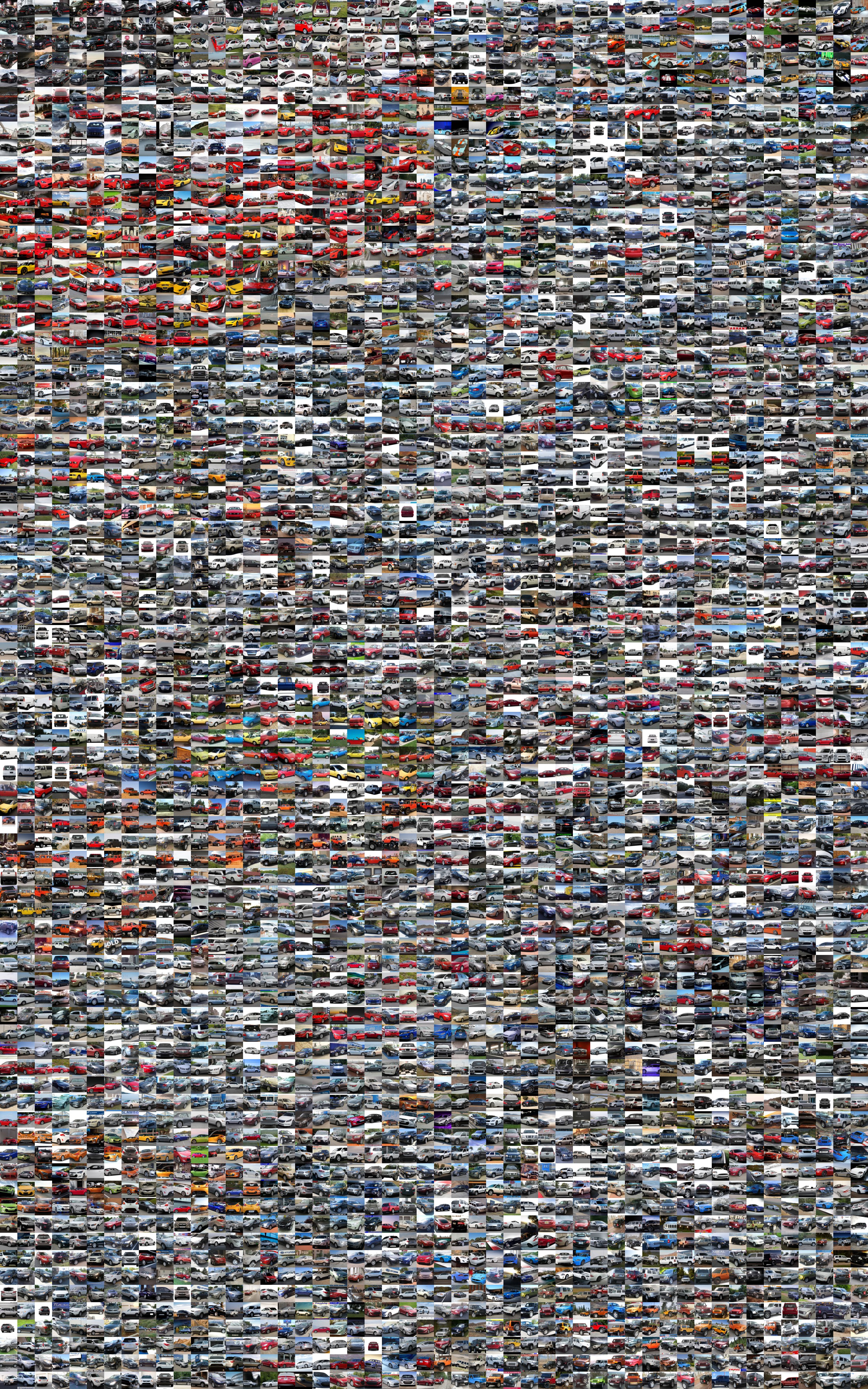}
\vspace{-2mm}
\caption{Cars-196 test subset (4000 images)}
\label{fig:cars_img}
\end{center}
\end{figure*}

\Cref{fig:cub_img,fig:cars_img} demonstrate actual images of the first 4000 samples of the evaluation split of CUB-200-2011 and Cars-196 datasets. We use the layout from \Cref{fig:cub_eval,fig:cars_eval} projected to a uniform 2D grid, preserving neighborhood relations of samples.

\clearpage

\clearpage

%%%%%%%%% REFERENCES
{\small
\bibliographystyle{ieee_fullname}
\bibliography{supp}
}